\documentclass[aos,preprint]{imsart}
\pdfoutput=1

\usepackage[OT1]{fontenc}
\usepackage[round]{natbib}
\usepackage[colorlinks,citecolor=blue,urlcolor=blue]{hyperref}
\usepackage{hypernat}

\usepackage{geometry}
\setattribute{journal}{name}{}

\usepackage{graphicx, subfigure, natbib, amsmath} 
\usepackage{algorithm, algorithmic, afterpage}
\usepackage{hyperref, enumerate, enumitem, ragged2e}
\DeclareMathOperator*{\argmin}{arg\,min} 
\usepackage{amsfonts}
\usepackage{bm}
\usepackage{tikz}
\usetikzlibrary{fit,positioning}
\newcommand{\mcPP}{\mathcal{PP}}

\newcommand{\mb}{\mathbf}

\newcommand{\mcX}{\mathcal{X}}
\newcommand{\R}{\mathbb{R}}
\newcommand{\distNorm}{\mathcal{N}}
\newcommand{\eq}{\begin{equation*}}
\newcommand{\en}{\end{equation*}}
\newcommand{\eqa}{\begin{eqnarray*}}
\newcommand{\ena}{\end{eqnarray*}}
\newcommand{\eqn}{\begin{equation}}
\newcommand{\enn}{\end{equation}}
\newcommand{\eqan}{\begin{eqnarray}}
\newcommand{\enan}{\end{eqnarray}}
\newcommand{\pmat}{\begin{pmatrix}}
\newcommand{\pman}{\end{pmatrix}}
\newcommand{\bitems}{\begin{itemize}}
\newcommand{\eitems}{\end{itemize}}
\newcommand{\benum}{\begin{enumerate}}
\newcommand{\eenum}{\end{enumerate}}
\newcommand{\bprf}{\begin{proof}}
\newcommand{\eprf}{\end{proof}}

\urldef\acmweb\url{http://people.seas.harvard.edu/~acm/}
\urldef\lbweb\url{http://people.fas.harvard.edu/~bornn/}
\urldef\rpaweb\url{http://people.seas.harvard.edu/~rpa/}
\urldef\kgweb\url{http://kirkgoldsberry.com/}

\begin{document}

\begin{frontmatter}
  \title{Factorized Point Process Intensities: A Spatial Analysis of Professional Basketball}
  \runtitle{Factorized Point Process Intensities}

  \begin{aug}
    \author{\fnms{Andrew} \snm{Miller}%
      \ead[label=e1]{acm@seas.harvard.edu}%
      \ead[label=u1,url]{http://people.seas.harvard.edu/\textasciitilde{}acm/}%
      \thanksref{t1}%
      },
    \author{\fnms{Luke} \snm{Bornn}%
      \ead[label=e2]{bornn@stat.harvard.edu}%
      \ead[label=u2,url]{http://people.fas.harvard.edu/\textasciitilde{}bornn/}%
      \thanksref{t2}%
      },
    \author{\fnms{Ryan} \snm{Adams}%
      \ead[label=e3]{rpa@seas.harvard.edu}%
      \ead[label=u3,url]{http://people.seas.harvard.edu/\textasciitilde{}rpa/}%
      \thanksref{t3}%
    }
    \and
    \author{\fnms{Kirk} \snm{Goldsberry}%
      \ead[label=e4]{kgoldsberry@fas.harvard.edu}%
      \ead[label=u4,url]{http://kirkgoldsberry.com/}%
      \thanksref{t4}%
    }
    \affiliation{Harvard University} 
    
    \thankstext{t1}{ \acmweb }
    \thankstext{t2}{ \lbweb }
    \thankstext{t3}{ \rpaweb }
    \thankstext{t4}{ \kgweb }

    \address{Andrew Miller\\
      School of Engineering and Applied Sciences\\
      Harvard University\\
      Cambridge, MA 02138, USA\\
      \printead{e1}\\
      \printead{u1}}

    \address{Luke Bornn \\
      Department of Statistics \\
      Harvard University \\
      Cambridge, MA 02138, USA\\
      \printead{e2}\\
      \printead{u2}}
    
    \address{Ryan Adams\\
      School of Engineering and Applied Sciences\\
      Harvard University\\
      Cambridge, MA 02138, USA\\
      \printead{e3}\\
      \printead{u3}}
    
    \address{Kirk Goldsberry\\
      Center for Geographic Analysis\\
      Harvard University\\
      Cambridge, MA 02138, USA\\
      \printead{e4}\\
      \printead{u4}}
    
    \runauthor{A.~Miller et al.}
    
  \end{aug}

\begin{abstract}
We develop a machine learning approach to represent and analyze the underlying spatial structure that governs shot selection among professional basketball players in the NBA.  
Typically, NBA players are discussed and compared in an heuristic, imprecise manner that relies on unmeasured intuitions about player behavior.  This makes it difficult to draw comparisons between players and make accurate player specific predictions.  
Modeling shot attempt data as a point process, we create a low dimensional representation of offensive player types in the NBA.  Using non-negative matrix factorization (NMF), an unsupervised dimensionality reduction technique, we show that a low-rank spatial decomposition summarizes the shooting habits of NBA players.  The spatial representations discovered by the algorithm correspond to intuitive descriptions of NBA player types, and can be used to model other spatial effects, such as shooting accuracy.
\end{abstract}

\end{frontmatter}

%
%

%
%
\section{Introduction}

The spatial locations of made and missed shot attempts in basketball are naturally modeled as a point process.  The Poisson process and its inhomogeneous variant are popular choices to model point data in spatial and temporal settings.  Inferring the latent intensity function, $\lambda(\cdot)$, is an effective way to characterize a Poisson process, and $\lambda(\cdot)$ itself is typically of interest.  Nonparametric methods to fit intensity functions are often desirable due to their flexibility and expressiveness, and have been explored at length \citep{cox1955, moller1998log, diggle2013statistical}.  Nonparametric intensity surfaces have been used in many applied settings, including density estimation \citep{adams-murray-mackay-2009c}, models for disease mapping \citep{benes2002bayesian}, and models of neural spiking \citep{cunningham2008inferring}.  

When data are related realizations of a Poisson process on the same space, we often seek the underlying structure that ties them together.  In this paper, we present an unsupervised approach to extract features from instances of point processes for which the intensity surfaces vary from realization to realization, but are constructed from a common library.  

The main contribution of this paper is an unsupervised method that finds a low dimensional representation of related point processes.  Focusing on the application of modeling basketball shot selection, we show that a matrix decomposition of Poisson process intensity surfaces can provide an interpretable feature space that parsimoniously describes the data.  
We examine the individual components of the matrix decomposition, as they provide an interesting quantitative summary of players' offensive tendencies.  
These summaries better characterize player types than any traditional categorization (e.g.~player position).  
One application of our method is personnel decisions.  Our representation can be used to select sets of players with diverse offensive tendencies. 
This representation is then leveraged in a latent variable model to visualize a player's field goal percentage as a function of location on the court.  


\subsection{Related Work}
Previously, \citet{adams-dahl-murray-2010a} developed a probabilistic matrix factorization method to predict score outcomes in NBA games.  Their method incorporates external covariate information, though they do not model spatial effects or individual players.  \citet{goldsberry} developed a framework for analyzing the defensive effect of NBA centers on shot frequency and shot efficiency.  Their analysis is restricted, however, to a subset of players in a small part of the court near the basket.

Libraries of spatial or temporal functions with a nonparametric prior have also been used to model neural data.  \citet{cunningham2009factor} develop the Gaussian process factor analysis model to discover latent `neural trajectories' in high dimensional neural time-series.  Though similar in spirit, our model includes a positivity constraint on the latent functions that fundamentally changes their behavior and interpretation. 


%
%
\section{Background}
\label{sec:background}
This section reviews the techniques used in our point process modeling method, including Gaussian processes (GPs), Poisson processes (PPs), log-Gaussian Cox processes (LGCPs) and non-negative matrix factorization (NMF).  

\subsection{Gaussian Processes}
A Gaussian process is a stochastic process whose sample path, ${f_1, f_2 \dots \in \R}$, is normally distributed.  GPs are frequently used as a probabilistic model over functions~${f: \mcX \rightarrow \R}$, where the realized value ${f_n \equiv f(x_n)}$ corresponds to a function evaluation at some point~${x_n \in \mcX}$.  The spatial covariance between two points in~$\mcX$ encode prior beliefs about the function~$f$; covariances can encode beliefs about a wide range of properties, including differentiability, smoothness, and periodicity.  

As a concrete example, imagine a smooth function~${f: \R^2 \rightarrow \R}$ for which we have observed a set of locations~${x_1, \dots, x_N}$ and values~${f_1, \dots, f_N}$.  We can model this `smooth' property by choosing a covariance function that results in smooth processes.  For instance, the squared exponential covariance function 
\eqan
  \text{cov}(f_i, f_j) = k(x_i, x_j) = \sigma^2 \exp \left( -\frac{1}{2}\frac{||x_i - x_j||^2}{\phi^2} \right)
  \label{eq:squared-exp}
\enan
assumes the function $f$ is infinitely differentiable, with marginal variation $\sigma^2$ and length-scale $\phi$, which controls the expected number of direction changes the function exhibits.  Because this covariance is strictly a function of the distance between two points in the space $\mcX$, the squared exponential covariance function is said to be stationary. 

We use this smoothness property to encode our inductive bias that shooting habits vary smoothly over the court space.  For a more thorough treatment of Gaussian processes, see \cite{rasmussen2006gaussian}.


\subsection{Poisson Processes}
A Poisson process is a completely spatially random point process on some space, $\mcX$, for which the number of points that end up in some set $A \subseteq \mcX$ is Poisson distributed.  We will use an inhomogeneous Poisson process on a domain $\mcX$.  That is, we will model the set of spatial points, $x_1, \dots, x_N$ with $x_n \in \mcX$, as a Poisson process with a non-negative intensity function~${\lambda(x): \mcX \rightarrow \R_+}$ (throughout this paper, $\R_+$ will indicate the union of the positive reals and zero).  This implies that for any set $A \subseteq \mcX$, the number of points that fall in $A$, $N_A$, will be Poisson distributed,
\eqan
  N_A &\sim& \textrm{Poiss}\left( \int_A \lambda(dA) \right).
\enan 
Furthermore, a Poisson process is `memoryless', meaning that $N_A$ is independent of $N_B$ for disjoint subsets $A$ and $B$.  We signify that a set of points~${\mb x \equiv \{ x_1, \dots, x_N \}}$ follows a Poisson process as 
\eqan
  \mb x &\sim& \mcPP(\lambda(\cdot)). 
\enan


One useful property of the Poisson process is the superposition theorem~\citep{kingman1992poisson}, which states that given a countable collection of independent Poisson processes $\mb x_1, \mb x_2, \dots$, each with measure $\lambda_1, \lambda_2, \dots$, their superposition is distributed as
\eqan
  \bigcup_{k=1}^\infty \mb x_k &\sim& \mcPP\left( \sum_{k=1}^\infty \lambda_k \right).
\enan
Furthermore, note that each intensity function $\lambda_k$ can be scaled by some non-negative factor and remain a valid intensity function.  The positive scalability of intensity functions and the superposition property of Poisson processes motivate the non-negative decomposition (Section~\ref{sec:nmf}) of a global Poisson process into simpler weighted sub-processes that can be shared between players.   

\subsection{Log-Gaussian Cox Processes}
A log-Gaussian Cox process (LGCP) is a doubly-stochastic Poisson process with a spatially varying intensity function modeled as an exponentiated GP
\eqan
  Z(\cdot) &\sim& \text{GP}(0, k(\cdot, \cdot) ) \\
  \lambda(\cdot) &\sim& \exp( Z(\cdot) ) \\
  x_1, \dots, x_N &\sim& \mcPP( \lambda(\cdot) )
\enan
where doubly-stochastic refers to two levels of randomness: the random function $Z(\cdot)$ and the random point process with intensity $\lambda(\cdot)$.  


\subsection{Non-Negative Matrix Factorization}
\label{sec:nmf}
Non-negative matrix factorization (NMF) is a dimensionality reduction technique that assumes some matrix $\mb \Lambda$ can be approximated by the product of two low-rank matrices
\eqan
  \mb \Lambda = \mb W \mb B
\enan
where the matrix $\mb \Lambda \in \R_+^{N \times V}$ is composed of $N$ data points of length $V$, the basis matrix $\mb B \in \R_+^{K \times V}$ is composed of $K$ basis vectors, and the weight matrix $\mb W \in \R_+^{N \times K}$ is composed of the $N$ non-negative weight vectors that scale and linearly combine the basis vectors to reconstruct $\mb \Lambda$.  Each vector can be reconstructed from the weights and the bases 
\eqan
  \mb \lambda_n = \sum_{k=1}^K W_{n,k} B_{k,:}. 
\enan
The optimal matrices $\mb W^*$ and $\mb B^*$ are determined by an optimization procedure that minimizes $\ell(\cdot, \cdot)$, a measure of reconstruction error or divergence between $\mb W \mb B$ and $\mb \Lambda$ with the constraint that all elements remain non-negative:
\eqan
  \mb W^*, \mb B^* &=& \argmin_{\mb W, \mb B \geq 0} \ell(\mb \Lambda, \mb W \mb B).
\enan
Different metrics will result in different procedures.  For arbitrary matrices $\mb X$ and $\mb Y$, one option is the squared Frobenius norm,
\eqan
  \ell_2(\mb X, \mb Y) &=& \sum_{i,j} (X_{ij} - Y_{ij})^2.
  \label{eq:frobenius}
\enan
Another choice is a matrix divergence metric
\begin{align}
  \ell_{\text{KL}}(\mb X, \mb Y) &= \sum_{i,j} X_{ij} \log \frac{X_{ij}}{Y_{ij}} - X_{ij} + Y_{ij}
  \label{eq:kl}
\end{align}
which reduces to the Kullback-Leibler (KL) divergence when interpreting matrices $\mb X$ and $\mb Y$ as discrete distributions, i.e.,~$\sum_{ij} X_{ij} = \sum_{ij} Y_{ij} = 1$~\citep{lee}.  Note that minimizing the divergence~$\ell_{\text{KL}}(\mb X, \mb Y)$ as a function of $\mb Y$ will yield a different result from optimizing over $\mb X$.  

The two loss functions lead to different properties of~$\mb W^*$ and~$\mb B^*$.  To understand their inherent differences, note that the $\text{KL}$ loss function includes a $\log$ ratio term.  This tends to disallow large \emph{ratios} between the original and reconstructed matrices, even in regions of low intensity.  In fact, regions of low intensity can contribute more to the loss function than regions of high intensity if the ratio between them is large enough.  The $\log$ ratio term is absent from the Frobenius loss function, which only disallows large \emph{differences}.  This tends to favor the reconstruction of regions of larger intensity, leading to more basis vectors focused on those regions.   

Due to the positivity constraint, the basis $\mb B^*$ tends to be disjoint, exhibiting a more `parts-based' decomposition than other, less constrained matrix factorization methods, such as PCA.  This is due to the restrictive property of the NMF decomposition that disallows negative bases to cancel out positive bases.  In practice, this restriction eliminates a large swath of `optimal' factorizations with negative basis/weight pairs, leaving a sparser and often more interpretable basis~\citep{lee1999learning}.  

%
%
\section{Data}
\label{sec:data}
\begin{figure}[t!]
\centering
\subfigure[points]{
  \includegraphics[height=.23\columnwidth, page=5]{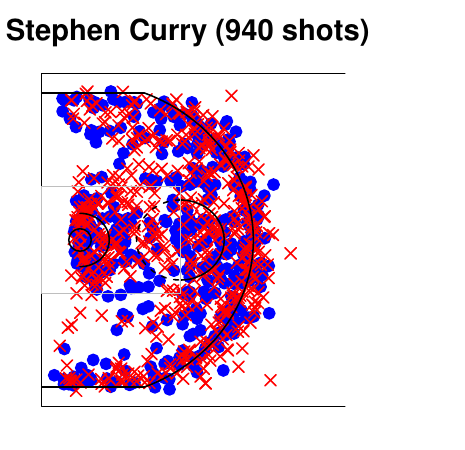}
  \includegraphics[height=.23\columnwidth, page=3]{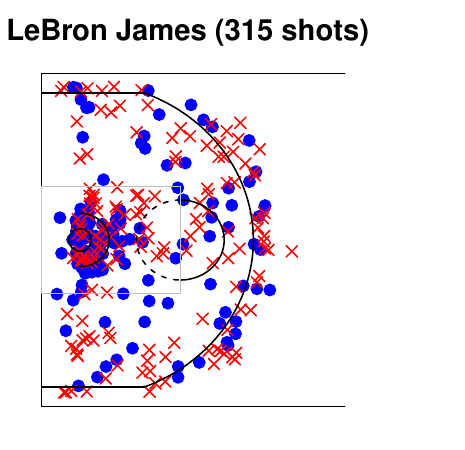}
  \label{fig:raw-data}
}
\subfigure[grid]{
  \includegraphics[height=.23\columnwidth, page=5]{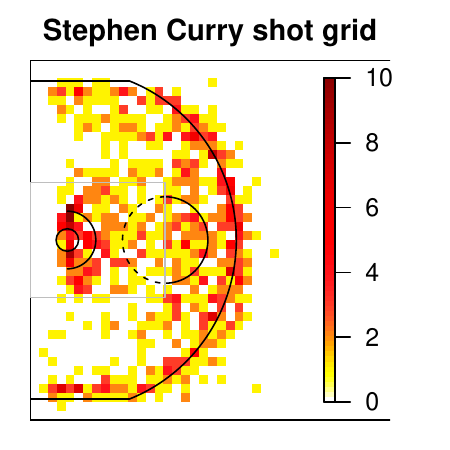}
  \includegraphics[height=.23\columnwidth, page=3]{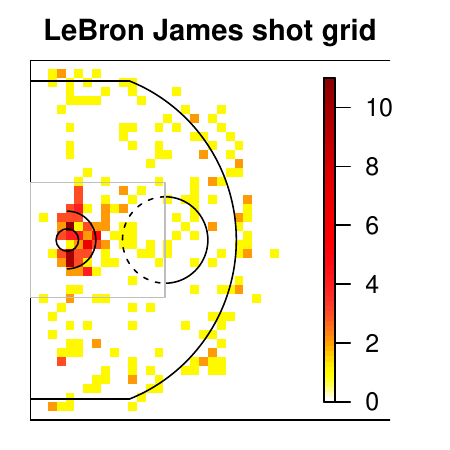}
}
\subfigure[LGCP]{
  \includegraphics[height=.23\columnwidth, page=5]{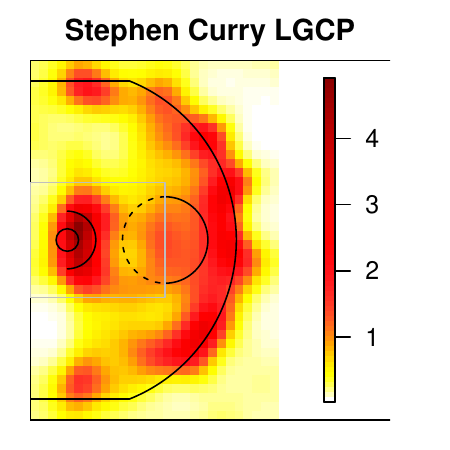}
  \includegraphics[height=.23\columnwidth, page=3]{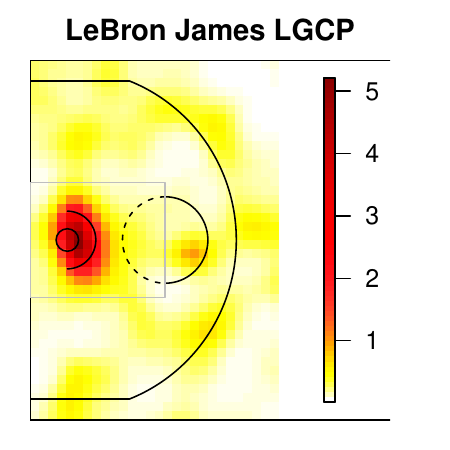}  
  \label{fig:lgcps}
}
\subfigure[LGCP-NMF]{
  \includegraphics[height=.23\columnwidth, page=5]{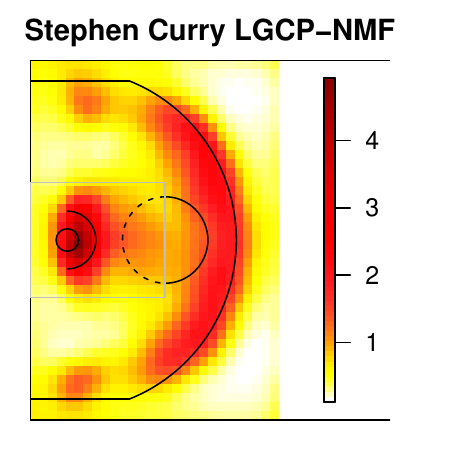}
  \includegraphics[height=.23\columnwidth, page=3]{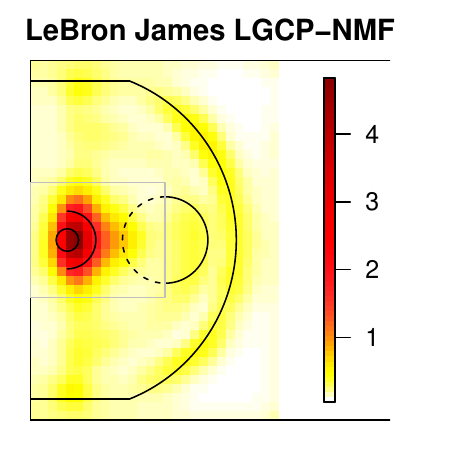}  
  \label{fig:nmf}
}
\caption{NBA player representations: (a) original point process data from two players, (b) discretized counts, (c) LGCP surfaces, and (d) NMF reconstructed surfaces ($K=10$).  Made and missed shots are represented as blue circles and red $\times$'s, respectively.  Some players have more data than others because only half of the stadiums had the tracking system in 2012-2013. }
\vspace{-1em}
\label{fig:first}
\end{figure}
\begin{figure}[t]
\centering   
\vspace{-1em}
\includegraphics[width=.28\columnwidth,page=5]{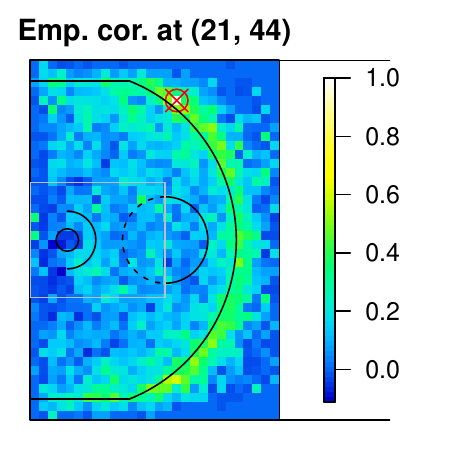}
\includegraphics[width=.28\columnwidth,page=3]{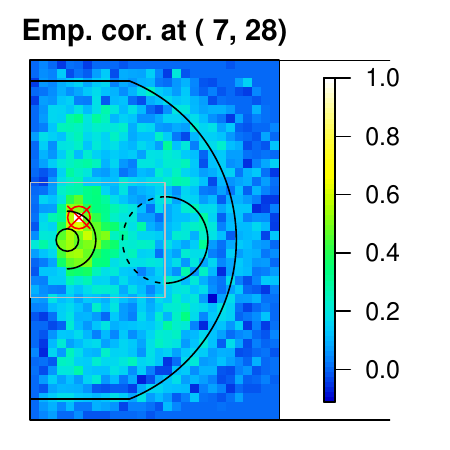} 
\vspace{-1em}
\caption{ Empirical spatial correlation in raw count data at two marked court locations.  These data exhibit non-stationary correlation patterns, particularly among three point shooters.  This suggests a modeling mechanism to handle the global correlation. }
\vspace{-1em}
\label{fig:emp_cov}
\end{figure}

Our data consist of made and missed field goal attempt locations from roughly half of the games in the 2012-2013 NBA regular season.  These data were collected by optical sensors as part of a program to introduce spatio-temporal information to basketball analytics.  We remove shooters with fewer than 50 field goal attempts, leaving a total of about 78,000 shots distributed among 335 unique NBA players.  

We model a player's shooting as a point process on the offensive half court, a 35 ft by 50 ft rectangle.  We will index players with~${n \in \{1, \dots, N\}}$, and we will refer to the set of each player's shot attempts as~${\mb x_n = \{ x_{n,1}, \dots, x_{n,M_n} \}}$, where $M_n$ is the number of shots taken by player $n$, and $x_{n,m} \in [0,35]\times[0,50]$.    

When discussing shot outcomes, we will use~${y_{n,m} \in \{0,1\}}$ to indicate that the~$n$th player's~$m$th shot was made (1) or missed (0).  Some raw data is graphically presented in Figure~\ref{fig:raw-data}.  Our goal is to find a parsimonious, yet expressive representation of an NBA basketball player's shooting habits.  

\subsection{A Note on Non-Stationarity}
As an exploratory data analysis step, we visualize the empirical spatial correlation of shot counts in a discretized space.  We discretize the court into $V$ tiles, and compute $\mb X$ such that $\mb X_{n,v} = |\{ x_{n,i} : x_{n,i} \in v\}|$, the number of shots by player $n$ in tile $v$.  The empirical correlation, depicted with respect to a few tiles in Figure~\ref{fig:emp_cov}, provides some intuition about the non-stationarity of the underlying intensity surfaces.  Long range correlations exist in clearly non-stationary patterns, and this inductive bias is not captured by a stationary LGCP that merely assumes a locally smooth surface.  This motivates the use of an additional method, such as NMF, to introduce global spatial patterns that attempt to learn this long range correlation.  


%
%
\section{Proposed Approach} 
\label{sec:approach}
Our method ties together the two ideas, LGCPs and NMF, to extract spatial patterns from NBA shooting data.  Given point process realizations for each of $N$ players, $\mb x_1, \dots, \mb x_N$, our procedure is
\vspace{-.25cm}
\begin{enumerate} \itemsep-2pt
\item Construct the count matrix $\mb X_{n,v} = \#$ shots by player $n$ in tile $v$ on a discretized court. 
\item Fit an intensity surface $\lambda_n = (\lambda_{n,1}, \dots, \lambda_{n,V})^T$ for each player $n$ over the discretized court (LGCP).
\item Construct the data matrix $\mb \Lambda = (\bar\lambda_1, \dots, \bar\lambda_N)^T$, where $\bar\lambda_n$ has been normalized to have unit volume.  
\item Find $\mb B, \mb W$ for some $K$ such that $\mb W \mb B \approx \mb \Lambda$, constraining all matrices to be non-negative (NMF).  
\end{enumerate}
\vspace{-.2cm}

This results in a spatial basis $\mb B$ and basis loadings for each individual player, $\mb w_n$.  Due to the superposition property of Poisson processes and the non-negativity of the basis and loadings, the basis vectors can be interpreted as sub-intensity functions, or archetypal intensities used to construct each individual player.  The linear weights for each player concisely summarize the spatial shooting habits of a player into a vector in $\R_+^K$.  

Though we have formulated a continuous model for conceptual simplicity, we discretize the court into~$V$ one-square-foot tiles to gain computational tractability in fitting the LGCP surfaces.  We expect this tile size to capture all interesting spatial variation.  Furthermore, the discretization maps each player into $\R_{+}^V$, providing the necessary input for NMF dimensionality reduction.  

\subsection{Fitting the LGCPs}
For each player's set of points, $\mb x_n$, the likelihood of the point process is discretely approximated as 
\vspace{-.15cm}
\eqan
  p(\mb x_n | \lambda_n(\cdot))   
    &\approx& \prod_{v=1}^{V} p(\mb X_{n,v} | \Delta A \lambda_{n,v} )
\enan
where, overloading notation, $\lambda_n(\cdot)$ is the exact intensity function, $\lambda_n$ is the discretized intensity function (vector), and $\Delta A$ is the area of each tile (implicitly one from now on).  This approximation comes from the completely spatially random property of the Poisson process, allowing us to treat each tile independently.  The probability of the count present in each tile is Poisson, with uniform intensity $\lambda_{n,v}$.  

Explicitly representing the Gaussian random field $\mb z_n$, the posterior is
\eqan
  p(\mb z_n | \mb x_n) 
    &\propto& p(\mb x_n | \mb z_n) p(\mb z_n) \\
    &=& \prod_{v=1}^{V} e^{-\lambda_{n,v}} \frac{\lambda_{n,v}^{\mb X_{n,v}}}{\mb X_{n,v}!} \distNorm( \mb z_n | 0, \mb K) \\
  \lambda_{n} &=& \exp( \mb z_n + z_0 )
\enan
where the prior over $\mb z_n$ is a mean zero normal with covariance~${\mb K_{v,u} = k(x_v, x_u)}$, determined by Equation \ref{eq:squared-exp}, and $z_0$ is a bias term that parameterizes the mean rate of the Poisson process.  Samples of the posterior~$p(\lambda_n | \mb x_n)$ can be constructed by transforming samples of~$\mb z_n | \mb x_n$.  To overcome the high correlation induced by the court's spatial structure, we employ elliptical slice sampling \citep{murray-adams-mackay-2010a} to approximate the posterior of $\lambda_n$ for each player, and subsequently store the posterior mean.

\subsection{NMF Optimization}
We now solve the optimization problem using techniques from \citet{lee} and \citet{brunet}, comparing the KL and Frobenius loss functions to highlight the difference between the resulting basis vectors.

%
%
\begin{table}[t]

\hskip2.2cm\includegraphics[width=.82\textwidth,page=1]{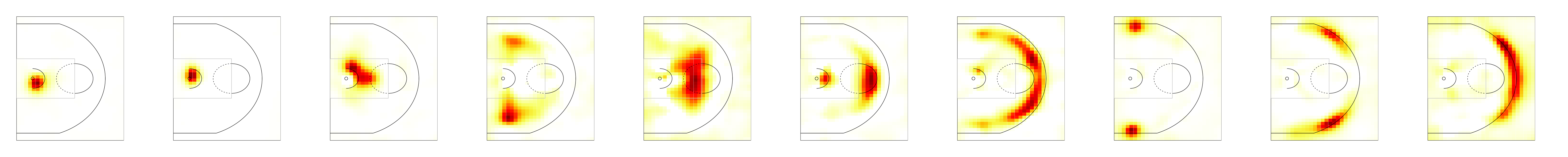} \\

\scalebox{0.9}{
\begin{tabular}{lp{3em}p{3em}p{3em}p{3em}p{3em}p{3em}p{3em}p{3em}p{3em}p{3em}}
  \hline
  \hline
LeBron James & \textcolor{red}{ 0.21 } & \textcolor{black}{ 0.16 } & \textcolor{black}{ 0.12 } & \textcolor{black}{ 0.09 } & \textcolor{black}{ 0.04 } & \textcolor{black}{ 0.07 } & \textcolor{black}{ 0.00 } & \textcolor{black}{ 0.07 } & \textcolor{black}{ 0.08 } & \textcolor{black}{ 0.17 } \\ 

  Brook Lopez & \textcolor{black}{ 0.06 } & \textcolor{red}{ 0.27 } & \textcolor{red}{ 0.43 } & \textcolor{black}{ 0.09 } & \textcolor{black}{ 0.01 } & \textcolor{black}{ 0.03 } & \textcolor{black}{ 0.08 } & \textcolor{black}{ 0.03 } & \textcolor{black}{ 0.00 } & \textcolor{black}{ 0.01 } \\ 
  Tyson Chandler & \textcolor{red}{ 0.26 } & \textcolor{red}{ 0.65 } & \textcolor{black}{ 0.03 } & \textcolor{black}{ 0.00 } & \textcolor{black}{ 0.01 } & \textcolor{black}{ 0.02 } & \textcolor{black}{ 0.01 } & \textcolor{black}{ 0.01 } & \textcolor{black}{ 0.02 } & \textcolor{black}{ 0.01 } \\ 
  Marc Gasol & \textcolor{black}{ 0.19 } & \textcolor{black}{ 0.02 } & \textcolor{black}{ 0.17 } & \textcolor{black}{ 0.01 } & \textcolor{red}{ 0.33 } & \textcolor{red}{ 0.25 } & \textcolor{black}{ 0.00 } & \textcolor{black}{ 0.01 } & \textcolor{black}{ 0.00 } & \textcolor{black}{ 0.03 } \\ 
  Tony Parker & \textcolor{black}{ 0.12 } & \textcolor{red}{ 0.22 } & \textcolor{black}{ 0.17 } & \textcolor{black}{ 0.07 } & \textcolor{red}{ 0.21 } & \textcolor{black}{ 0.07 } & \textcolor{black}{ 0.08 } & \textcolor{black}{ 0.06 } & \textcolor{black}{ 0.00 } & \textcolor{black}{ 0.00 } \\ 
  Kyrie Irving & \textcolor{black}{ 0.13 } & \textcolor{black}{ 0.10 } & \textcolor{black}{ 0.09 } & \textcolor{black}{ 0.13 } & \textcolor{black}{ 0.16 } & \textcolor{black}{ 0.02 } & \textcolor{black}{ 0.13 } & \textcolor{black}{ 0.00 } & \textcolor{black}{ 0.10 } & \textcolor{black}{ 0.14 } \\ 
  Stephen Curry & \textcolor{black}{ 0.08 } & \textcolor{black}{ 0.03 } & \textcolor{black}{ 0.07 } & \textcolor{black}{ 0.01 } & \textcolor{black}{ 0.10 } & \textcolor{black}{ 0.08 } & \textcolor{red}{ 0.22 } & \textcolor{black}{ 0.05 } & \textcolor{black}{ 0.10 } & \textcolor{red}{ 0.24 } \\ 
  James Harden & \textcolor{red}{ 0.34 } & \textcolor{black}{ 0.00 } & \textcolor{black}{ 0.11 } & \textcolor{black}{ 0.00 } & \textcolor{black}{ 0.03 } & \textcolor{black}{ 0.02 } & \textcolor{black}{ 0.13 } & \textcolor{black}{ 0.00 } & \textcolor{black}{ 0.11 } & \textcolor{red}{ 0.26 } \\ 
  Steve Novak & \textcolor{black}{ 0.00 } & \textcolor{black}{ 0.01 } & \textcolor{black}{ 0.00 } & \textcolor{black}{ 0.02 } & \textcolor{black}{ 0.00 } & \textcolor{black}{ 0.00 } & \textcolor{black}{ 0.01 } & \textcolor{red}{ 0.27 } & \textcolor{red}{ 0.35 } & \textcolor{red}{ 0.34 } \\ 
   \hline
\end{tabular}
}

\caption{ Normalized player weights for each basis.  The first three columns correspond to close-range shots, the next four correspond to mid-range shots, while the last three correspond to three-point shots.  Larger values are highlighted, revealing the general `type' of shooter each player is.  The weights themselves match intuition about players shooting habits (e.g.~three-point specialist or mid-range shooter), while more exactly quantifying them. }
\label{tab:weights}
\end{table}

\begin{figure}[!]
\vspace{0em}
\centering    
\subfigure[Corner threes]{\label{fig:basis1}\includegraphics[width=.24\columnwidth,page=1]{"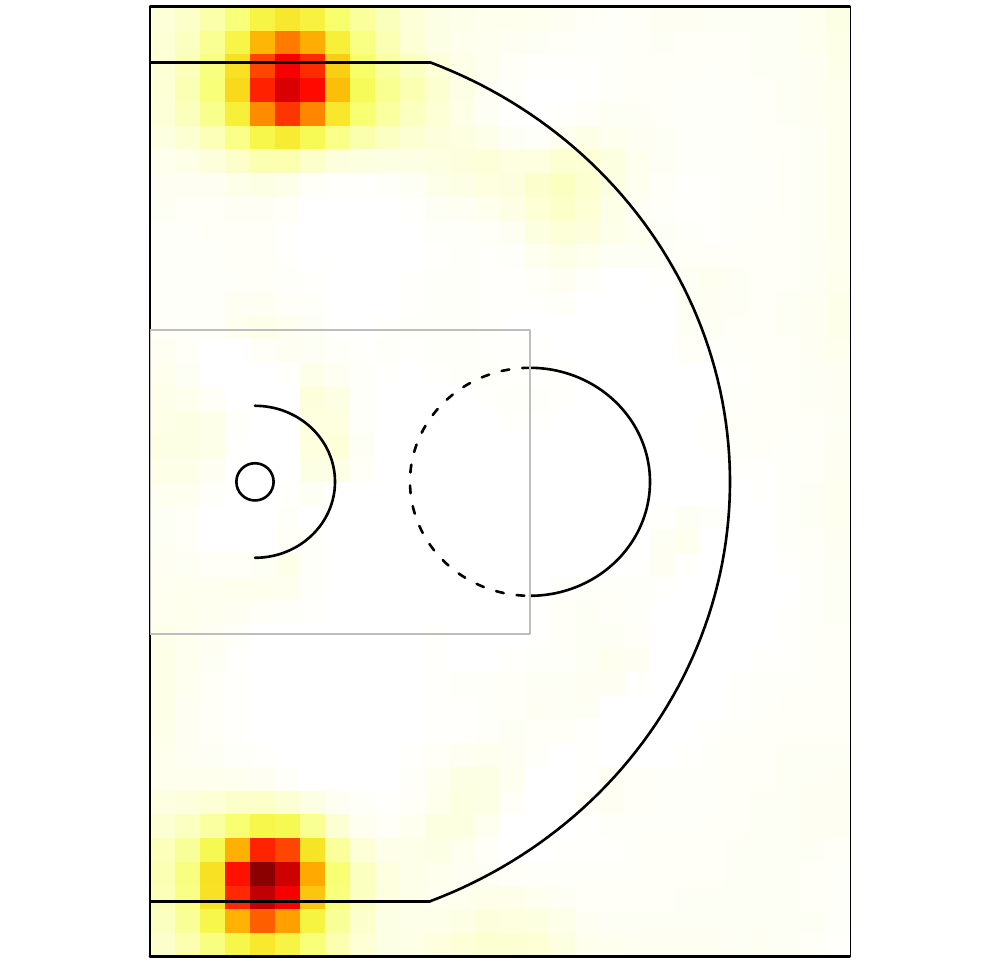}}
\subfigure[Wing threes]{\label{fig:basis2}\includegraphics[width=.24\columnwidth, page=4]{"figs/k_10.pdf}}
\subfigure[Top of key threes]{\label{fig:basis3}\includegraphics[width=.24\columnwidth, page=10]{"figs/k_10.pdf}}
\subfigure[Long two-pointers]{\label{fig:basis4}\includegraphics[width=.24\columnwidth, page=8]{"figs/k_10.pdf}}
\caption{ A sample of basis vectors (surfaces) discovered by LGCP-NMF for $K=10$.  Each basis surface is the normalized intensity function of a particular shot type, and players' shooting habits are a weighted combination of these shot types.  Conditioned on certain shot type (e.g. corner three), the intensity function acts as a density over shot locations, where red indicates likely locations.  }
\label{fig:basis}
\end{figure}

%
%
\begin{figure}[h!]
\centering    
\subfigure[LGCP-NMF (KL)]{\label{fig:brunet}\includegraphics[width=\columnwidth]{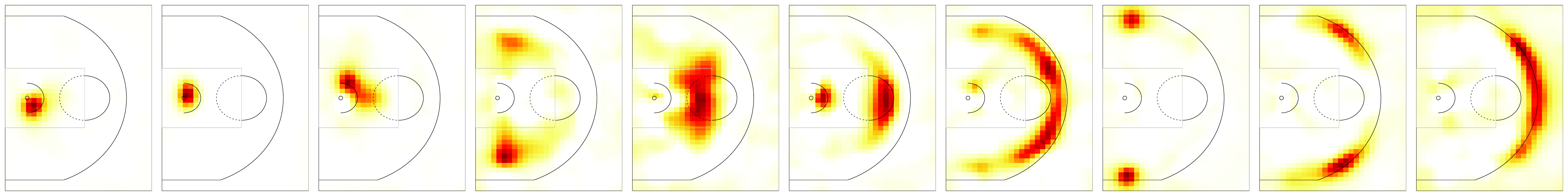}}\vspace{-.5em}
\subfigure[LGCP-NMF (Frobenius)]{\label{fig:lee}\includegraphics[width=\columnwidth]{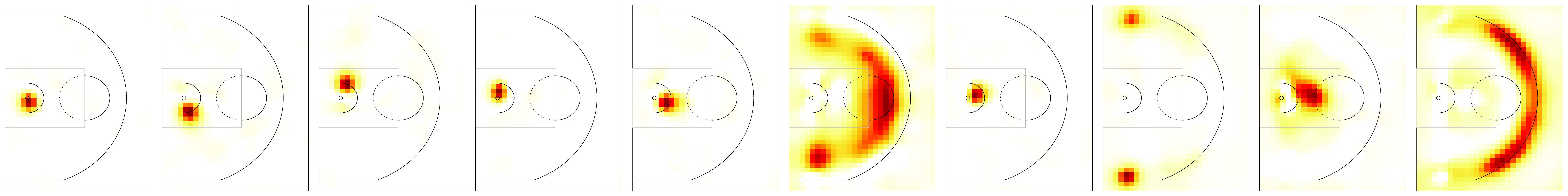}}\vspace{-.5em}
\subfigure[Direct NMF (KL)]{\label{fig:raw}\includegraphics[width=\columnwidth]{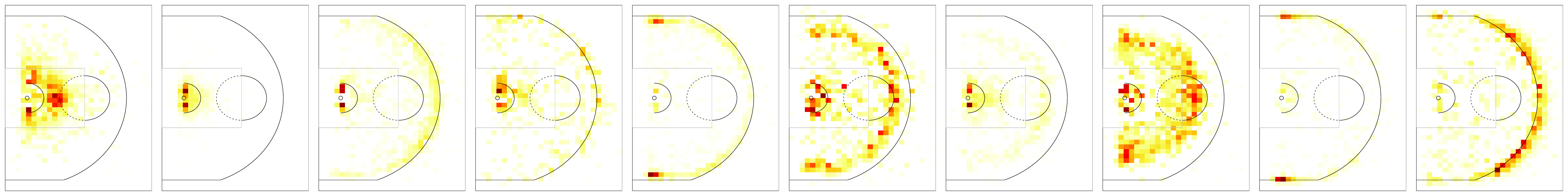}}\vspace{-.5em}
\subfigure[LGCP-PCA]{\label{fig:pca}\includegraphics[width=\columnwidth]{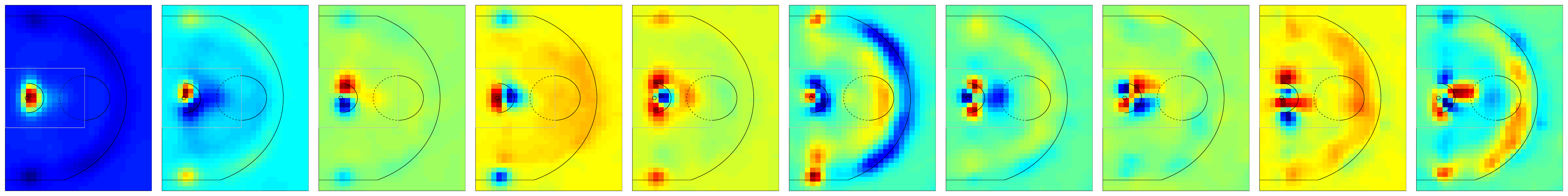}}\vspace{-.5em}
\caption{Visual comparison of the basis resulting from various approaches to dimensionality reduction.  The top two bases result from LGCP-NMF with the KL (top) and Frobenius (second) loss functions.  The third row is the NMF basis applied to raw counts (no spatial continuity).   The bottom row is the result of PCA applied to the LGCP intensity functions.  LGCP-PCA fundamentally differs due to the negativity of the basis surfaces. Best viewed in color. }
\vspace{-1em}
\label{fig:nmf-methods}
\end{figure}
\afterpage{\newpage}

\subsection{Alternative Approaches}
With the goal of discovering the shared structure among the collection of point processes, we can proceed in a few alternative directions. For instance, one could hand-select a spatial basis and directly fit weights for each individual point process, modeling the intensity as a weighted combination of these bases.  However, this leads to multiple restrictions: firstly, choosing the spatial bases to cover the court is a highly subjective task (though, there are situations where it would be desirable to have such control); secondly, these bases  are unlikely to match the natural symmetries of the basketball court.  In contrast, modeling the intensities with LGCP-NMF uncovers the natural symmetries of the game without user guidance.

%

Another approach would be to directly factorize the raw shot count matrix $\mb X$.  However, this method ignores spatial information, and essentially models the intensity surface as a set of $V$ independent parameters.  Empirically, this method yields a poorer, more degenerate basis, which can be seen in Figure~\ref{fig:raw}.  Furthermore, this is far less numerically stable, and jitter must be added to entries of $\mb \Lambda$ for convergence.  
Finally, another reasonable approach would apply PCA directly to the discretized LGCP intensity matrix $\mb \Lambda$, though as Figure \ref{fig:pca} demonstrates, the resulting mixed-sign decomposition leads to an unintuitive and visually uninterpretable basis.

\section{Results}
\label{sec:results}
We graphically depict our point process data, LGCP representation, and LGCP-NMF reconstruction in Figure \ref{fig:first} for ${K=10}$. There is wide variation in shot selection among NBA players - some shooters specialize in certain types of shots, whereas others will shoot from many locations on the court.  

Our method discovers basis vectors that correspond to visually interpretable shot types.  Similar to the parts-based decomposition of human faces that NMF discovers in \citet{lee1999learning}, LGCP-NMF discovers a shots-based decomposition of NBA players. 

Setting ${K=10}$ and using the KL-based loss function, we display the resulting basis vectors in Figure~\ref{fig:basis}.  One basis corresponds to corner three-point shots \ref{fig:basis1}, while another corresponds to wing three-point shots \ref{fig:basis2}, and yet another to top of the key three point shots \ref{fig:basis3}.  A comparison between KL and Frobenius loss functions can be found in Figure~\ref{fig:nmf-methods}.  

Furthermore, the player specific basis weights provide a concise characterization of their offensive habits.  The weight $w_{n,k}$ can be interpreted as the amount player $n$ takes shot type $k$, which quantifies intuitions about player behavior. Table~\ref{tab:weights} compares normalized weights between a selection of players.  

Empirically, the KL-based NMF decomposition results in a more spatially diverse basis, where the Frobenius-based decomposition focuses on the region of high intensity near the basket at the expense of the rest of the court.  This can be seen by comparing Figure~\ref{fig:brunet} (KL) to Figure~\ref{fig:lee} (Frobenius).  

We also compare the two LGCP-NMF decompositions to the NMF decomposition done directly on the matrix of counts, $\mb X$.  The results in Figure~\ref{fig:raw} show a set of sparse basis vectors that are spatially unstructured.   And lastly, we depict the PCA decomposition of the LGCP matrix $\mb \Lambda$ in Figure~\ref{fig:pca}.  This yields the most colorful decomposition because the basis vectors and player weights are unconstrained real numbers.  This renders the basis vectors uninterpretable as intensity functions.  Upon visual inspection, the corner three-point `feature' that is salient in the LGCP-NMF decompositions appears in five separate PCA vectors, some positive, some negative.  This is the cancelation phenomenon that NMF avoids.  

\begin{figure}[t!]
\centering    
\vspace{-1.1em}
\includegraphics[width=.8\columnwidth]{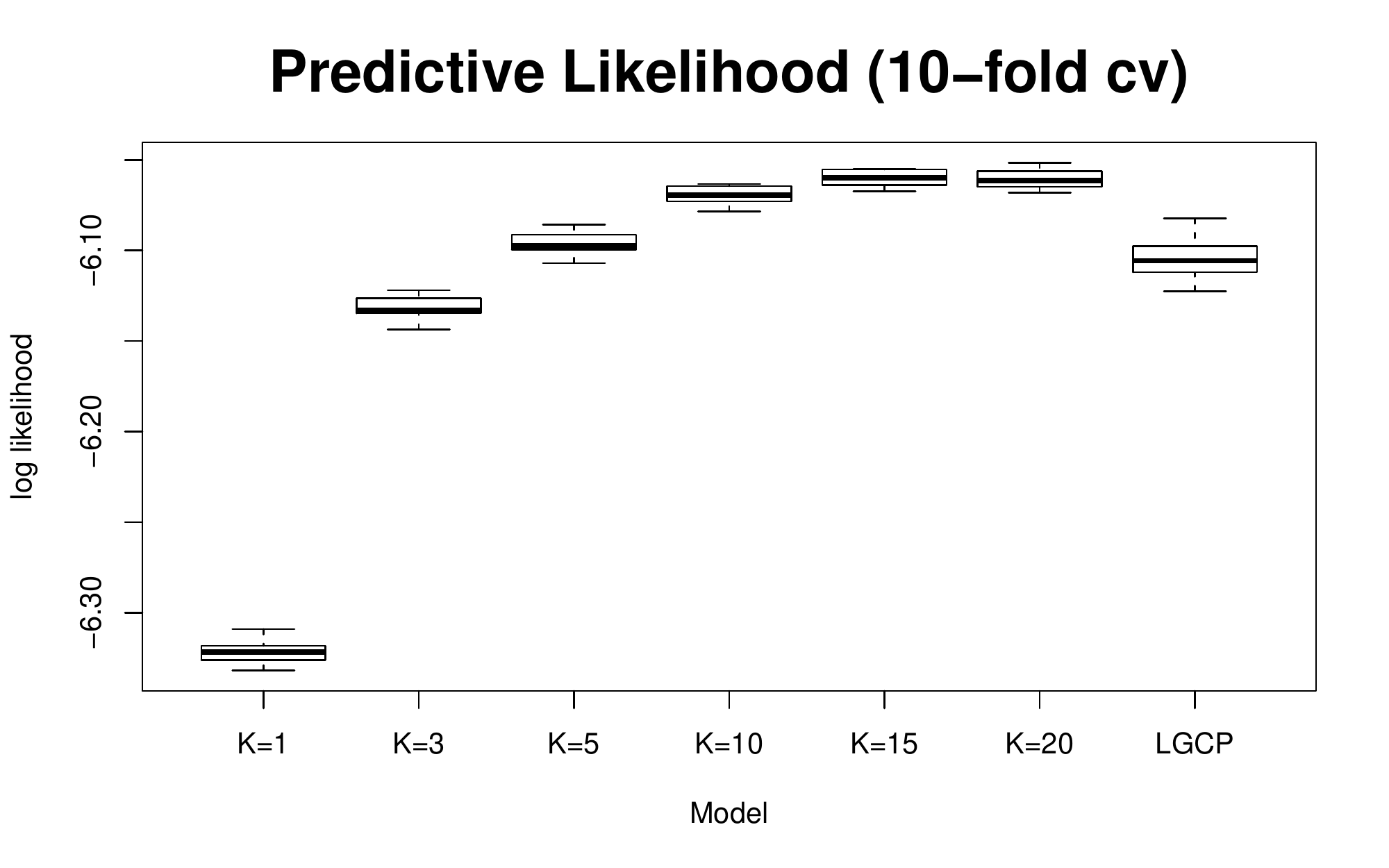}
\vspace{-1.2em}
\caption{ Average player test data log likelihoods for LGCP-NMF varying $K$ and independent LGCP.  For each fold, we held out 10\% of each player's shots, fit independent LGCPs and ran NMF (using the KL-based loss function) for varying $K$.  We display the average (across players) test log likelihood above.  The predictive performance of our representation improves upon the high dimensional independent LGCPs, showing the importance of pooling information across players. }
\vspace{-.7em}
\label{fig:test-points}
\end{figure}

We compare the fit of the low rank NMF reconstructions and the original LGCPs on held out test data in Figure~\ref{fig:test-points}.  The NMF decomposition achieves superior predictive performance over the original independent LGCPs in addition to its compressed representation and interpretable basis.  

%
%
\section{From Shooting Frequency to Efficiency}
\label{sec:application}
Unadjusted field goal percentage, or the probability a player makes an attempted shot, is a statistic of interest when evaluating player value.  This statistic, however, is spatially uninformed, and washes away important variation due to shooting circumstances. 

Leveraging the vocabulary of shot types provided by the basis vectors, we model a player's field goal percentage for each of the shot types.  We decompose a player's field goal percentage into a weighted combination of $K$ basis field goal percentages, which provides a higher resolution summary of an offensive player's skills.  Our aim is to estimate the probability of a made shot for each point in the offensive half court \emph{for each individual player}. 

\subsection{Latent variable model}
For player $n$, we model each shot event as 
\begin{align*}
  k_{n,i} &\sim \text{Mult}( \bar w_{n,:} ) && \text{ shot type } \\
  \color{blue}{x_{n,i}| k_{n,i}}  &\sim \text{Mult}(\bar B_{k_{n,i}}) && \text{ location } \\
  \color{red}{y_{n,i} | k_{n,i}}  &\sim \text{Bern}( \text{logit}^{-1}( \beta_{n,k_{n,i}} ) ) && \text{ outcome }
\end{align*}
where $\bar B_k \equiv B_k / \sum_{k'} B_{k'}$ is the normalized basis, and the player weights $\bar w_{n,k}$ are adjusted to reflect the total mass of each unnormalized basis.  NMF does not constrain each basis vector to a certain value, so the volume of each basis vector is a meaningful quantity that corresponds to how common a shot type is.  We transfer this information into the weights by setting  
\begin{align*}
  \bar w_{n,k} &= w_{n,k} \sum_v B_k(v).  && \text{ adjusted basis loadings }
\end{align*}
We do not directly observe the shot type, $k$, only the shot location $x_{n,i}$. Omitting $n$ and $i$ to simplify notation, we can compute the the predictive distribution
\begin{align*}
  p(y | x) &= \sum_{k=1}^K {\color{red}p(y | k)} p(k | x) \\
           &= \sum_{z=1}^K {\color{red}p(y | k)} \frac{ {\color{blue}p(x | k)} p(k) }{ \sum_{k'} {\color{blue}p(x | k')} p(k') }
\end{align*}
where the outcome distribution is red and the location distribution is blue for clarity. 



The shot type decomposition given by $\mb B$ provides a natural way to share information between shooters to reduce the variance in our estimated surfaces.  We hierarchically model player probability parameters $\beta_{n,k}$ with respect to each shot type.  The prior over parameters is
\begin{align*}
    \beta_{0,k} &\sim \distNorm(0,\sigma_0^2) && \text{ diffuse global prior } \\
    \sigma_{k}^2  &\sim \text{Inv-Gamma}(a, b)  && \text{ basis variance } \\
    \beta_{n,k} &\sim \distNorm(\beta_{0,k}, \sigma_k^2) && \text{ player/basis params}
\end{align*}
where the global means, $\beta_{0,k}$, and variances, $\sigma_k^2$, are given diffuse priors, $\sigma_0^2 = 100$, and $a = b =.1$.
The goal of this hierarchical prior structure is to share information between players about a particular shot type.  Furthermore, it will shrink players with low sample sizes to the global mean.  Some consequences of these modeling decisions will be discussed in Section~\ref{discussion}.  

\subsection{Inference}
Gibbs sampling is performed to draw posterior samples of the $\beta$ and $\sigma^2$ parameters.  To draw posterior samples of $\beta | \sigma^2, y$, we use elliptical slice sampling to exploit the normal prior placed on $\beta$.  We can draw samples of $\sigma^2 | \beta, y$ directly due to conjugacy. 

\subsection{Results}
\begin{figure}[t!]
\vspace{-1.3em}
\centering
\subfigure[global mean]{\label{fig:global-mean} \includegraphics[width=.3\columnwidth, page=1]{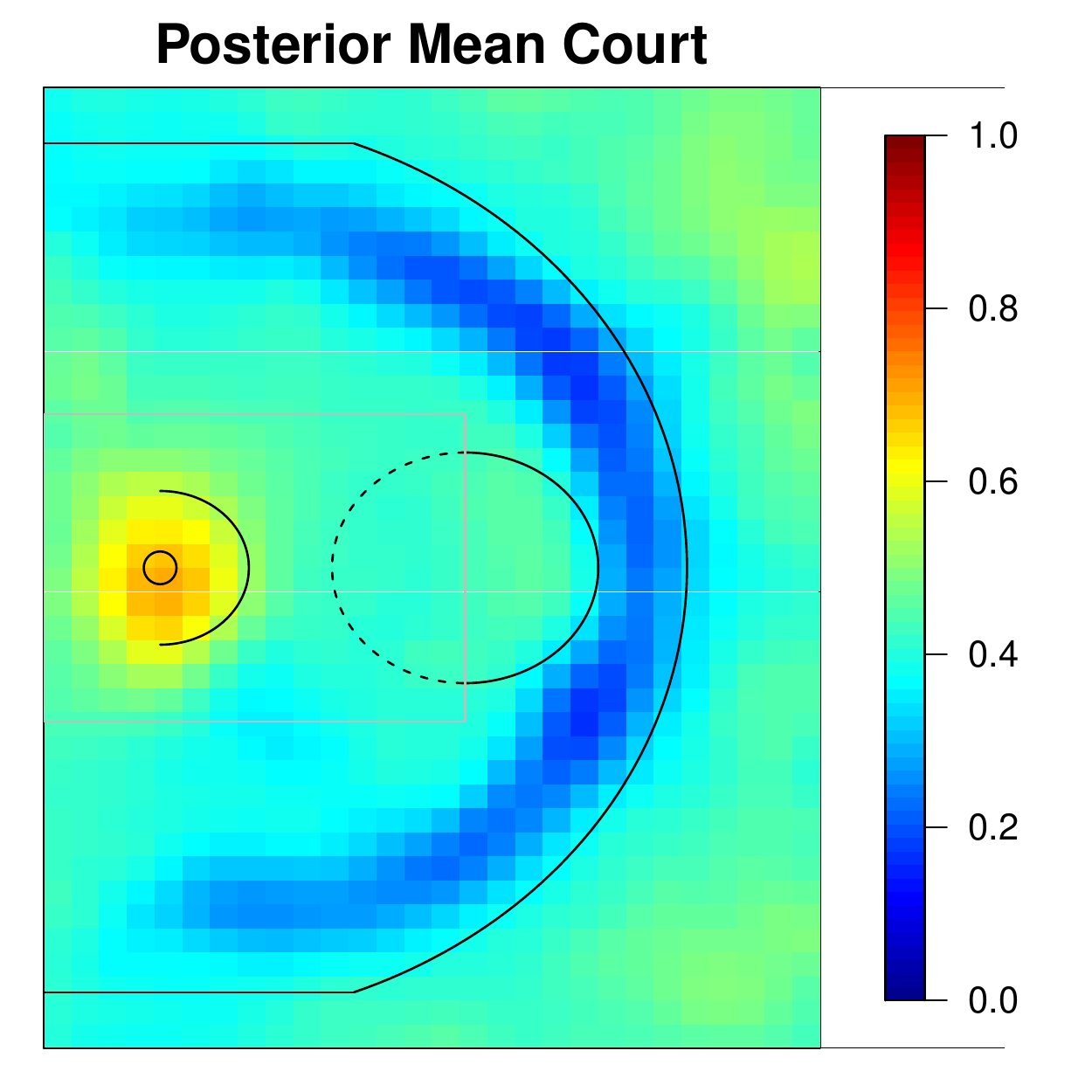} }
\subfigure[posterior uncertainty]{ \includegraphics[width=.3\columnwidth, page=2]{figs/ind_lvm_hier_global.pdf} }
\subfigure[]{ \includegraphics[width=.3\columnwidth, page=1]{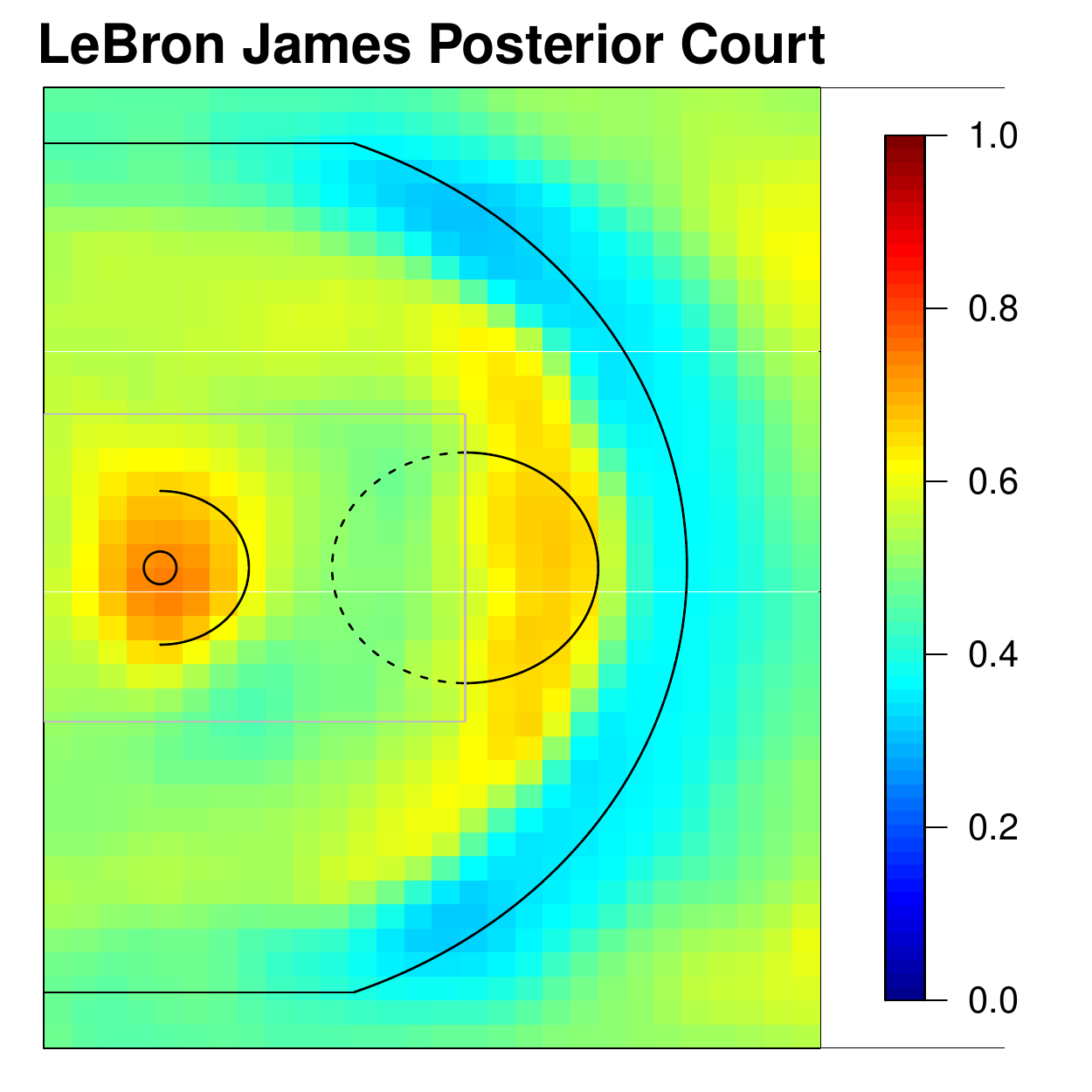} }
\subfigure[]{ \includegraphics[width=.3\columnwidth, page=121]{figs/ind_lvm_hier_players.pdf} }
\subfigure[]{ \includegraphics[width=.3\columnwidth, page=76]{figs/ind_lvm_hier_players.pdf} }
\subfigure[]{ \includegraphics[width=.3\columnwidth, page=91]{figs/ind_lvm_hier_players.pdf} }
\vspace{-1em}
\caption{ (a) Global efficiency surface and (b) posterior uncertainty.  (c-f) Spatial efficiency for a selection of players.  Red indicates the highest field goal percentage and dark blue represents the lowest.  Novak and Curry are known for their 3-point shooting, whereas James and Irving are known for efficiency near the basket.  } 
\vspace{-1.2em}
\label{fig:efficiency}
\end{figure}
We visualize the global mean field goal percentage surface, corresponding parameters to $\beta_{0,k}$ in Figure~\ref{fig:global-mean}.  Beside it, we show one standard deviation of posterior uncertainty in the mean surface. Below the global mean, we show a few examples of individual player field goal percentage surfaces.  These visualizations allow us to compare players' efficiency with respect to regions of the court.  For instance, our fit suggests that both Kyrie Irving and Steve Novak are below average from basis 4, the baseline jump shot, whereas Stephen Curry is an above average corner three point shooter.  This is valuable information for a defending player.  More details about player field goal percentage surfaces and player parameter fits are available in the supplemental material.


\section{Discussion}
\label{discussion}
We have presented a method that models related point processes using a constrained matrix decomposition of independently fit intensity surfaces.  Our representation provides an accurate low dimensional summary of shooting habits and an intuitive basis that corresponds to shot types recognizable by basketball fans and coaches.  After visualizing this basis and discussing some of its properties as a quantification of player habits, we then used the decomposition to form interpretable estimates of a spatially shooting efficiency. 

We see a few directions for future work.  Due to the relationship between KL-based NMF and some fully generative latent variable models, including the probabilistic latent semantic model \citep{ding2008equivalence} and latent Dirichlet allocation \citep{blei2003latent}, we are interested in jointly modeling the point process and intensity surface decomposition in a fully generative model.  This spatially informed LDA would model the non-stationary spatial structure the data exhibit within each non-negative basis surface, opening the door for a richer parameterization of offensive shooting habits that could include defensive effects.

Furthermore, jointly modeling spatial field goal percentage and intensity can capture the correlation between player skill and shooting habits.  Common intuition that players will take more shots from locations where they have more accuracy is missed in our treatment, yet modeling this effect may yield a more accurate characterization of a player's habits and ability.    

 
\section*{Acknowledgments} 
The authors would like to acknowledge the Harvard XY Hoops group, including Alex Franks, Alex D'Amour, Ryan Grossman, and Dan Cervone.  We also acknowledge the HIPS lab and several referees for helpful suggestions and discussion, and STATS LLC for providing the data.  To compare various NMF optimization procedures, the authors used the \texttt{r} package \texttt{NMF} \citep{r-nmf}.  

\bibliographystyle{plainnat}
\bibliography{refs}

\end{document}